\pdfoutput=1

\documentclass{article}

\usepackage{microtype}
\usepackage{graphicx}
\usepackage{booktabs}

\usepackage[accepted]{icml2019}

\usepackage[]{amsmath}
\usepackage{macros}
\usepackage{subcaption}
\usepackage{xcolor}
\usepackage{bm}
\usepackage{mathtools}
\usepackage{mwe}
\usepackage{adjustbox}
\usepackage{array}
\usepackage{enumitem}
\usepackage{multirow}
\usepackage{natbib}

\setcitestyle{numbers}
\setcitestyle{square}
\setcitestyle{comma}

\definecolor{phicolor}{HTML}{0074D9}
\definecolor{gammacolor}{HTML}{FF851B}

\newcommand{\ourtitle}{Cubic-Spline Flows}

\icmltitlerunning{\ourtitle}

\usepackage{hyperref}
\usepackage{cleveref}

\begin{document}

\twocolumn[
\icmltitle{\ourtitle}

\icmlsetsymbol{equal}{*}

\begin{icmlauthorlist}
\icmlauthor{Conor Durkan}{equal,edi}
\icmlauthor{Artur Bekasov}{equal,edi}
\icmlauthor{Iain Murray}{edi}
\icmlauthor{George Papamakarios}{edi}
\end{icmlauthorlist}

\icmlaffiliation{edi}{School of Informatics, University of Edinburgh, United Kingdom}

\icmlcorrespondingauthor{Conor Durkan}{conor.durkan@ed.ac.uk}
\icmlcorrespondingauthor{Artur Bekasov}{artur.bekasov@ed.ac.uk}

\icmlkeywords{Normalizing flows, Density estimation, Generative modeling}

\vskip 0.3in
]

\printAffiliationsAndNotice{\icmlEqualContribution}

% Abstract
\begin{abstract}
A normalizing flow models a complex probability density as an invertible transformation of a simple density. The invertibility means that we can evaluate densities and generate samples from a flow. In practice, autoregressive flow-based models are slow to invert, making either density estimation or sample generation slow. Flows based on coupling transforms are fast for both tasks, but have previously performed less well at density estimation than autoregressive flows. We stack a new coupling transform, based on monotonic cubic splines, with LU-decomposed linear layers. The resulting \emph{cubic-spline flow} retains an exact one-pass inverse, can be used to generate high-quality images, and closes the gap with autoregressive flows on a suite of density-estimation tasks.
\end{abstract}

% Sections
\section{Introduction}

Normalizing flows are flexible probabilistic models of continuous data. A \emph{normalizing flow} models data $\bfx$ as the output of an invertible smooth transformation $\bff$ of noise $\bfu$:
\begin{equation}
    \bfx = \bff\roundbr{\bfu}
    \quad\text{where}\quad
    \bfu\sim\pi\roundbr{\bfu}.
    \label{eq:flow_sampling}
\end{equation}
Typically, the noise distribution $\pi\roundbr{\bfu}$ is taken to be simple (a standard normal is a common choice), whereas the transformation $\bff$ and its inverse $\bff^{-1}$ are implemented by an invertible neural network.
The probability density of $\bfx$ under the flow is obtained by a change of variables:
\begin{equation}
    p\roundbr{\bfx} = \pi\roundbr{\bff^{-1}\roundbr{\bfx}}\,\vertbr{\det\roundbr{\frac{\partial \bff^{-1}}{\partial\bfx}}}.
    \label{eq:flow_density}
\end{equation}
 Given training data $\curlybr{\bfx_1,\ldots, \bfx_N}$, the flow is trained with backpropagation, by maximizing the total log likelihood $\sum_{n=1}^N{\log p\roundbr{\bfx_n}}$ with respect to the parameters of the transformation $\bff$.
Normalizing flows have been successfully used for density estimation \citep{Dinh:2017:rnvp, Papamakarios:2017:maf}, variational inference \citep{Kingma:2016:iaf, Rezende:2015:flows, Berg:2018:sylvester}, image, audio and video generation \citep{Kim:2018:FloWaveNet, Kingma:2018:glow, Kumar:2019:VideoFlow, Prenger:2018:WaveGlow}, and likelihood-free inference \citep{Papamakarios:2019:snl}.

In practice, the Jacobian determinant of $\bff$ must be tractable, so that the density $p\roundbr{\bfx}$ can be computed efficiently. Several invertible transformations with a tractable Jacobian determinant have been proposed, including coupling transforms \citep{Dinh:2015:nice, Dinh:2017:rnvp}, autoregressive transforms \citep{DeCao:2019:bnaf, Huang:2018:naf, Kingma:2016:iaf, Papamakarios:2017:maf}, invertible convolutions \citep{Hoogeboom:2019:emerging, Kingma:2018:glow}, transformations based on the matrix-determinant lemma \citep{Rezende:2015:flows, Berg:2018:sylvester}, and continuous transformations based on differential equations \citep{Chen:2018:neural_odes, Grathwohl:2018:ffjord}.

Even though the above transformations are invertible in \hbox{theory}, they are not always efficiently invertible in practice. For instance, the affine autoregressive transforms used by MAF \citep{Papamakarios:2017:maf} and IAF \citep{Kingma:2016:iaf} are $D$ times slower to invert than to evaluate, where $D$ is the dimensionality of $\bfx$. Even worse, inverting the non-affine transformations used by NAF~\citep{Huang:2018:naf} and block-NAF \citep{DeCao:2019:bnaf} would require numerical optimization. In practice, the forward transformation $\bff$ is needed for generating data, whereas the inverse transformation $\bff^{-1}$ is needed for computing $p\roundbr{\bfx}$; for a flow to be applicable in both situations, both directions need to be fast.

Transformations that are equally fast to invert and evaluate do exist. Affine coupling transforms, used by Real NVP \citep{Dinh:2017:rnvp} and Glow \citep{Kingma:2018:glow}, only require a single neural-network pass in either direction; however, more flexible autoregressive transforms have obtained better density-estimation results. More flexible coupling transforms have been proposed, based on piecewise-quadratic splines \citep{muller2018neuralimportancesampling}, although these have not previously been evaluated as density estimators. Continuous flows such as Neural ODEs \citep{Chen:2018:neural_odes} and FFJORD \citep{Grathwohl:2018:ffjord} are also equally fast in both directions, but they require numerically integrating a differential equation in each direction, which can be slower than a single neural-network pass.

In this paper we introduce \emph{cubic-spline flows}, which, like Real NVP and Glow, only require a single neural-network pass in either the forward or the inverse direction, but in practice are as flexible as state-of-the-art autoregressive flows. Cubic-spline flows alternate between non-affine coupling transforms that use flexible monotonic cubic splines to transform their input, and LU-decomposed linear layers that combine their input across dimensions. Experimentally, we show that cubic-spline flows match or exceed state-of-the-art performance in density estimation, and that they are capable of generating high-quality images with only a single neural-network pass.\looseness=-1

\begin{figure}[t]
	\centering
	\begin{subfigure}{0.49\columnwidth}
    	\centering
	    \includegraphics[width=\columnwidth]{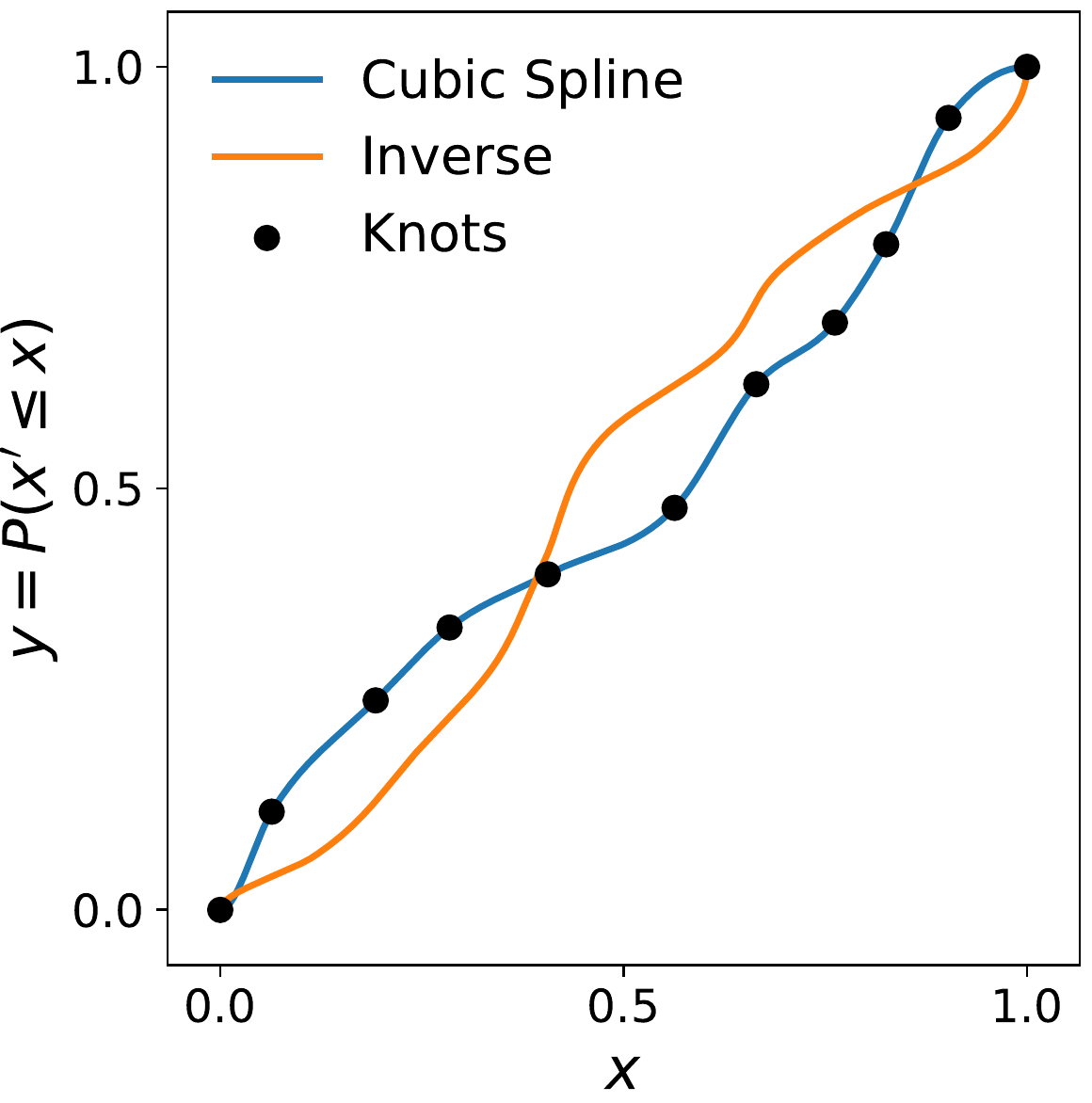}
	\end{subfigure}
	\begin{subfigure}{0.49\columnwidth}
	    \centering
	    \includegraphics[width=\columnwidth]{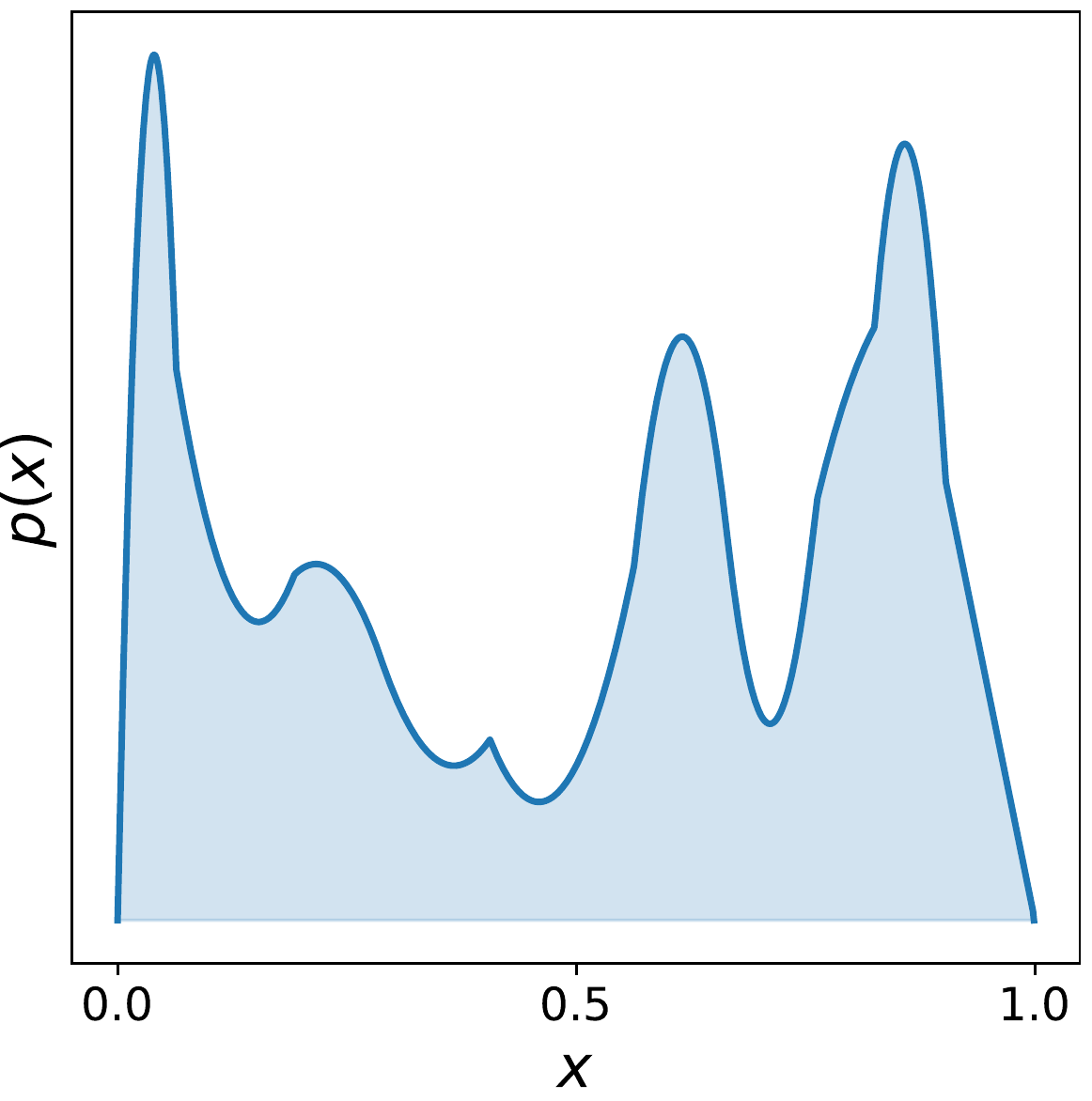}
	\end{subfigure}
	\caption{\small \textbf{Left}: Illustration of a random monotonic cubic spline with $ K = 10 $ bins which maps $ [0, 1] $ to $ [0, 1] $. Monotonic splines act as drop-in replacements for affine or additive transformations in coupling layers. \textbf{Right}: Since the cubic spline can be interpreted as a cumulative distribution function, its derivative defines a probability density function which is piecewise quadratic.}
	\label{fig:cubic-spline}
\end{figure}

\section{Method}
\label{sec:method}
\subsection{Coupling transforms}
A \textit{coupling transform} $\bm{\phi}$ maps an input $ \bfx $ to an output $ \bfy $ in the following way: 
\begin{enumerate}[leftmargin=*, topsep=0pt, itemsep=0pt]
    \item Split the input $ \bfx $ into two parts, $ \bfx = \squarebr{\bfx_{1:d-1}, \bfx_{d:D}} $.
    \item Compute parameters $ \bftheta = \text{NN}(\bfx_{1:d-1}) $, where NN is an arbitrary neural network. 
    \item Compute $ y_i = g_{\bftheta_i}(x_i) $ for $i=d:D$, where $g_{\bftheta_i}\!$ is an invertible function parameterized by $ \bftheta_i $.
    \item Set $ \bfy_{1:d-1} = \bfx_{1:d-1} $, and return $ \bfy = \squarebr{\bfy_{1:d-1}, \bfy_{d:D}} $.
\end{enumerate}
The Jacobian matrix of a coupling transform is lower triangular, since $ \bfy_{d:D} $ is given by transforming $ \bfx_{d:D} $ elementwise as a function of $ \bfx_{1:d-1} $, and $ \bfy_{1:d-1} $ is equal to $ \bfx_{1:d-1} $.
Thus, the Jacobian determinant of the coupling transform $\bm{\phi}$ is
\begin{equation}
\det\roundbr{\frac{\partial\bm{\phi}}{\partial\bfx}}   = \prod_{i=d}^{D} \frac{\partial g_{\bftheta_i}}{\partial x_i}.
\end{equation}
Coupling transforms solve two important problems for normalizing flows: first, they have a tractable Jacobian determinant; second,
they can be inverted exactly in a single pass. 
The inverse of a coupling transform can be easily computed by running steps 1--4 above, this time inputting $ \bfy $, and using $g_{\bftheta_i}^{-1} $ to compute $ \bfx_{d:D} $ in step 3.

\subsection{Monotonic-spline coupling transforms}\label{ssec:spline_transforms}
Typically, the function $ g_{\bftheta_i} \!$ takes the form of an \textit{affine} or \textit{additive} transformation for computational ease. The affine transformation is given by:
\begin{equation}
    g_{\bftheta_i}(x_i) = \alpha_i x_i + \beta_i
    \quad\text{where}\quad
    \bftheta_i = \curlybr{\alpha_{i}, \beta_{i}}.
\end{equation}
The additive transformation corresponds to the special case $\alpha_i=1$.
Both the affine and the additive transformation are easy to invert, but they lack flexibility.

Nonetheless, we are free to choose any function $g_{\bftheta_i} \!$ as long as it is differentiable and we can easily invert it. 
Recently, \citet{muller2018neuralimportancesampling} proposed a powerful generalization of the above forms, based on monotonic piecewise polynomials. 
The idea is to restrict the input domain of $g_{\bftheta_i}\! $ between $0$ and $1$, partition the input domain into $K$ bins, and define $g_{\bftheta_i}\! $ to be a simple polynomial segment within each bin (see \cref{fig:cubic-spline}). 
\citet{muller2018neuralimportancesampling} restrict themselves to monotonically-increasing linear and quadratic polynomial segments, whose coefficients are parameterized by $\bftheta_i$. 
Moreover, the polynomial segments are restricted to match at the boundaries so that  $g_{\bftheta_i} \!$ is continuous. Functions of this form are known as \emph{polynomial splines}.

\subsection{Monotonic cubic-spline coupling transforms}\label{ssec:cubic_spline_transforms}
Inspired by this direction of research, we introduce \emph{cubic-spline flows}, a natural extension to the framework of \citet{muller2018neuralimportancesampling}.
We propose to implement $g_{\bftheta_i}\!$ as a \emph{monotonic cubic spline} \cite{fritsch1980monotone}, where each segment is defined by a monotonically-increasing cubic polynomial (\cref{fig:cubic-spline}).
Cubic splines are well-known across science, finding use in many applications, and their theory is well-explored \cite{durrleman1989cubicsplinesb, hou1978cubicsplinesa}.

We employ \emph{Steffen's method} \cite{steffen1990simple} to parameterize a  monotonically-increasing cubic spline that maps $[0, 1]$ to $[0, 1]$. 
Given a fixed number of bins $ K $, Steffen's method takes \mbox{$K+1$} coordinates $\curlybr{(x_k, y_k)}_{k=0}^{K}$, known as \emph{knots}, and two values for the derivatives at inputs $0$ and $1$. Then, the method constructs a continuously-differentiable cubic spline that passes through the knots, has the given boundary derivatives, and is monotonic on each segment (for details, see \cref{sec:spline_parameterization}).
To ensure the spline is monotonically-increasing and maps $[0, 1]$ to $[0, 1]$, we require:
\begin{align}
&(x_0, y_0) = (0, 0), \quad (x_K, y_K) = (1, 1), \\
&x_{k} < x_{k+1} \,\text{ and }\, y_{k} < y_{k+1} \,\, \text{ for }k=0:K-1.
\end{align}
Our implementation of the \emph{cubic-spline coupling transform} is as follows:
\begin{enumerate}[leftmargin=*, topsep=0pt, itemsep=0pt]
    \item A neural network NN takes $\bfx_{1:d-1}$ and outputs an unconstrained parameter vector $ \bftheta_{i} $ of length $2K+2$ for each $i=d:D$.
    \item Vector $ \bftheta_{i} $ is partitioned as $ \bftheta_{i} =\squarebr{\bftheta_{i}^{w},\bftheta_{i}^{h}, \bftheta_{i}^{d}}$, where $\bftheta_{i}^{w}$ and $\bftheta_{i}^{h}$ have length $K$, and $\bftheta_{i}^{d}$ has length $2$.
    \item Vectors $\bftheta_{i}^{w}$ and $\bftheta_{i}^{h}$ are each passed through a softmax; the outputs are interpreted as the widths and heights of the $ K $ bins, which must be positive and sum to $1$.
    \item A cumulative sum of the $K$ bin widths and heights yields the $K+1$ knots $\curlybr{(x_k, y_k)}_{k=0}^{K}$. Vector $\bftheta_{i}^{d}$ is interpreted as the two boundary derivatives.
\end{enumerate}

Evaluating a cubic spline at location $x$ requires finding the bin in which $x$ lies, which can be done efficiently with binary search, since the bins are sorted. 
The Jacobian determinant of the cubic-spline coupling transform can be computed in closed form as a product of quadratic-polynomial derivatives (see \cref{sec:spline_derivative}). 
For the inverse, we need to compute the roots of a cubic polynomial whose constant term depends on the desired value to invert, which can be done analytically.
However, in early experiments, we found that na\"ive root finding using trigonometric and hyperbolic methods was unstable for certain values of the coefficients.
To address this issue, we follow a modified version, provided by \citet{christoph2016howtosolve}, of the scheme outlined by \citet{blinn2007solve}, which stabilizes many of the potentially difficult cases (see \cref{sec:spline_inverse}). 

Unlike the affine and additive transformations which have limited flexibility, 
a monotonic spline can approximate any continuous monotonic function arbitrarily well, given sufficiently many bins. Yet, cubic-spline coupling transforms have a closed-form Jacobian determinant and can be inverted in one pass. Finally, our parameterization of cubic splines using Steffen's method is fully differentiable, so the transform can be trained by gradient methods.

In addition to transforming $ \bfx_{d:D} $ by monotonic cubic splines whose parameters are a function of $ \bfx_{1:d-1} $, we introduce a set of monotonic cubic splines which act elementwise on $ \bfx_{1:d-1} $, whose parameters are optimized directly with stochastic gradient descent. This means that our coupling layer transforms all input variables at once, rather than being restricted to a subset, as follows:
\begin{align}
    \bftheta_{1:d-1} &= \text{Trainable parameters}\\ 
    \bftheta_{d:D} &= \text{NN}(\bfx_{1:d-1})\label{eq:coupling_nn}\\ 
    y_i &= g_{\bftheta_i}(x_i) \quad\text{for each } i=1:D.
\end{align}

\subsection{Linear transforms with an LU decomposition}\label{ssec:lu_transform}
To ensure all input variables can interact with each other, it is common to permute the elements of intermediate layers in a normalizing flow. 
Permutation is an invertible linear transformation, with absolute determinant equal to $1$. 
\citet{Kingma:2018:glow} generalize this approach using an efficient parameterization of a linear transformation, demonstrating improvements on a range of image tasks. 
In particular, they parameterize the \textit{lower-upper} or \textit{LU decomposition} of a square matrix as   $\bfW = \bfP \bfL \bfU$,
where $ \bfP $ is a fixed permutation matrix, $ \bfL $ is lower-triangular with ones on the diagonal, and $ \bfU $ is upper triangular. 
Written in this form, the determinant of $ \bfW $ can be calculated in $\bigo{D}$ time as the product of the diagonal elements of $\bfU$. Also, by parameterizing the diagonal elements of $\bfU$ to be positive, $ \bfW $ is guaranteed to be invertible. Given that we have the LU factorization of $\bfW$, we can efficiently apply the inverse transformation, and sample from the model in one pass.

\subsection{Cubic-spline flows}
We propose a general-purpose flow based on alternating LU-decomposed linear layers and monotonic cubic-spline coupling transforms. 
Since the splines map $ [0, 1] $ to $ [0, 1] $, we apply a sigmoid, and its inverse, the logit, before and after each coupling layer, so the overall transform is compatible with unconstrained data and linear layers.
We clip the inputs of the logit to $[10^{-6}, 1 \!-\! 10^{-6}]$ to prevent saturation with 32-bit floating-point precision.
The function is no longer strictly reversible for all inputs due to this clipping.

Like Real NVP or Glow, cubic-spline flows can represent either a transformation from data to noise, or from noise to data. 
In both cases, the transformation requires only a single pass of the neural network defining the flow. 
Overall, our proposed cubic-spline flow resembles a traditional feed-forward neural network architecture, alternating between linear transformations and piecewise nonlinearities, while retaining exact invertibility.

\newcommand{\plotsizeplane}{0.84in}
\renewcommand\arraystretch{0.5}
\begin{figure}{}
    	\centering
    	\setlength\tabcolsep{1pt} 
    	\begin{tabular}{ccc}
    	    Training data & Flow density & Flow samples \\
        	\includegraphics[width=\plotsizeplane]{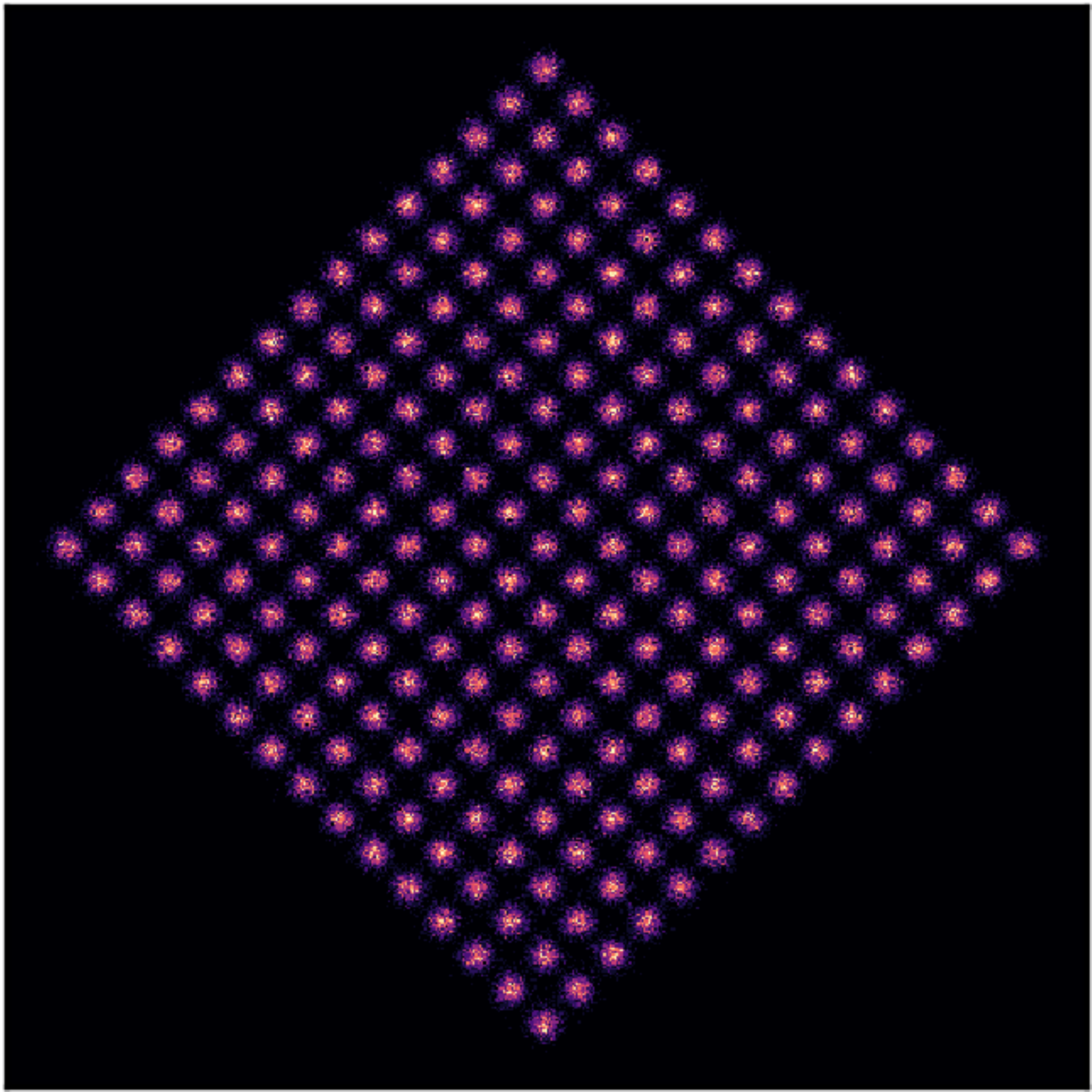} &
        	\includegraphics[width=\plotsizeplane]{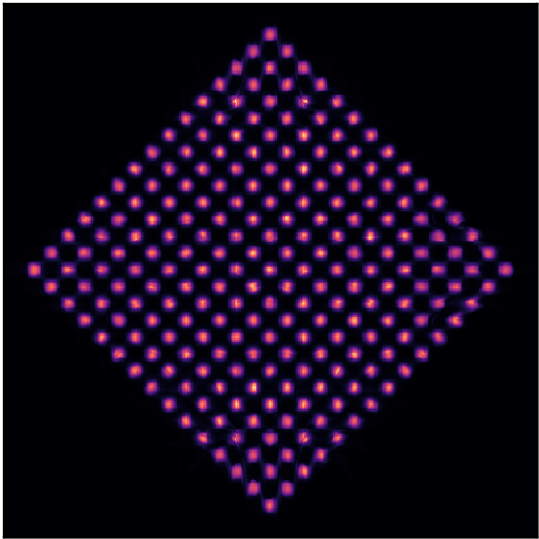} &
        	\includegraphics[width=\plotsizeplane]{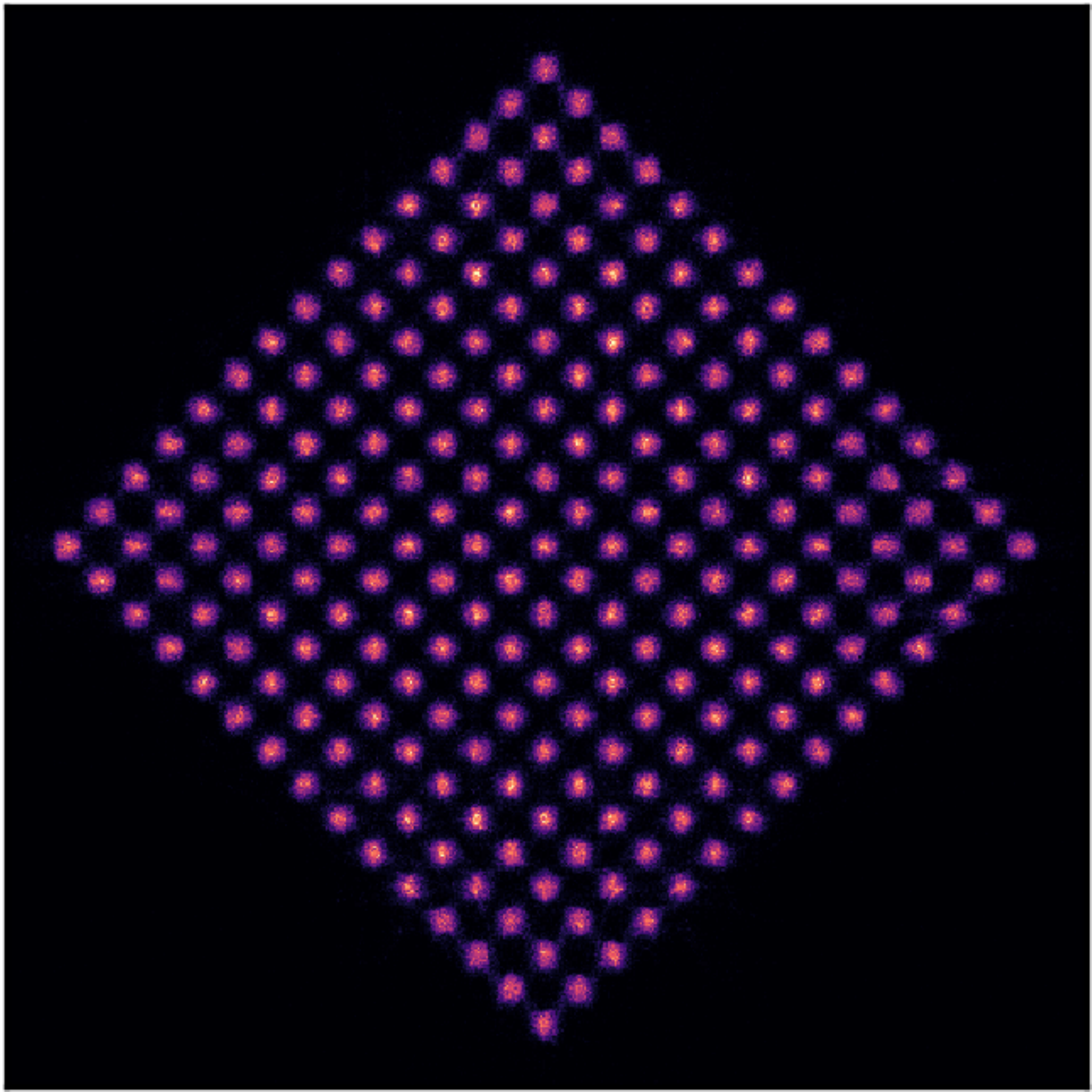} 
        	\\
        	\includegraphics[width=\plotsizeplane]{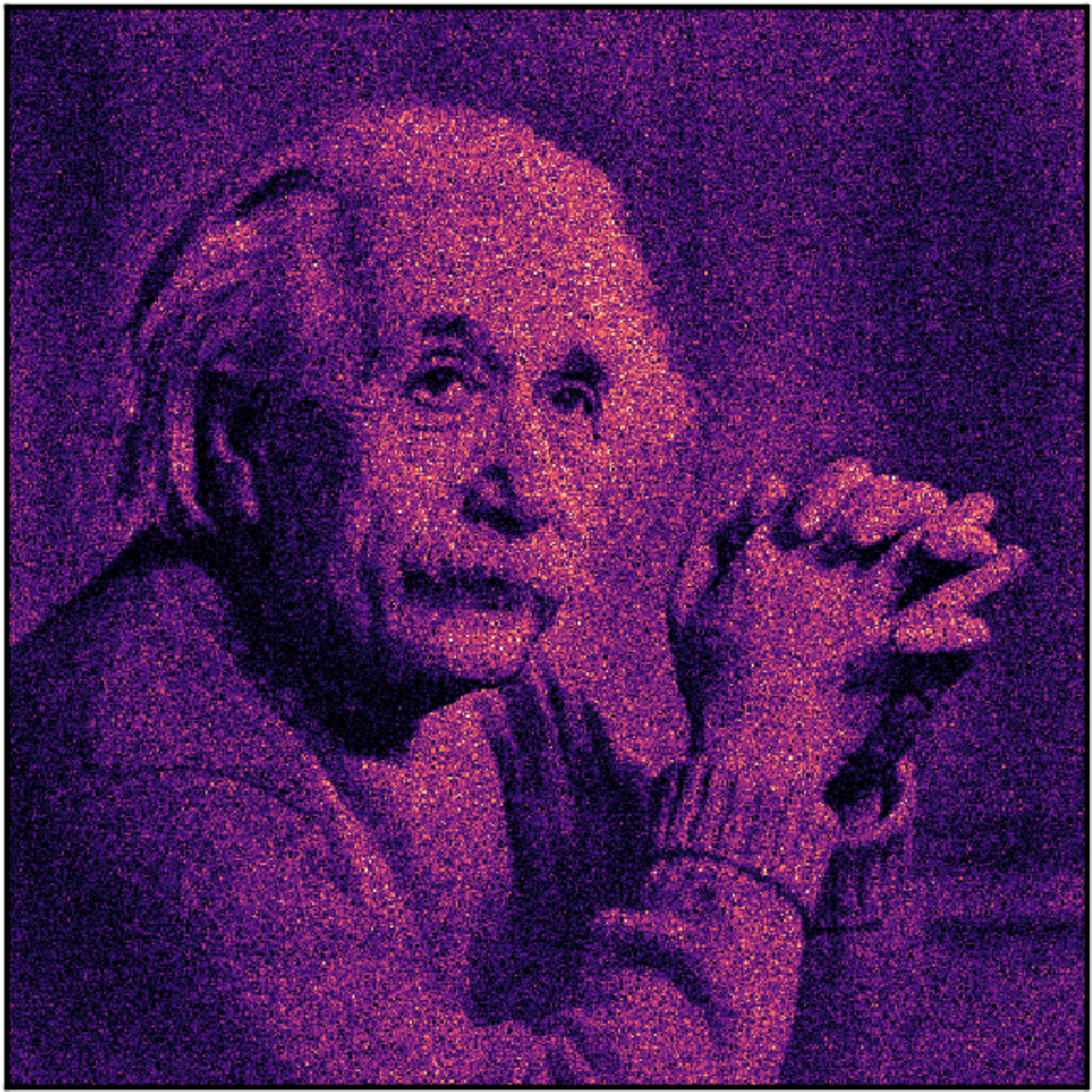} &
        	\includegraphics[width=\plotsizeplane]{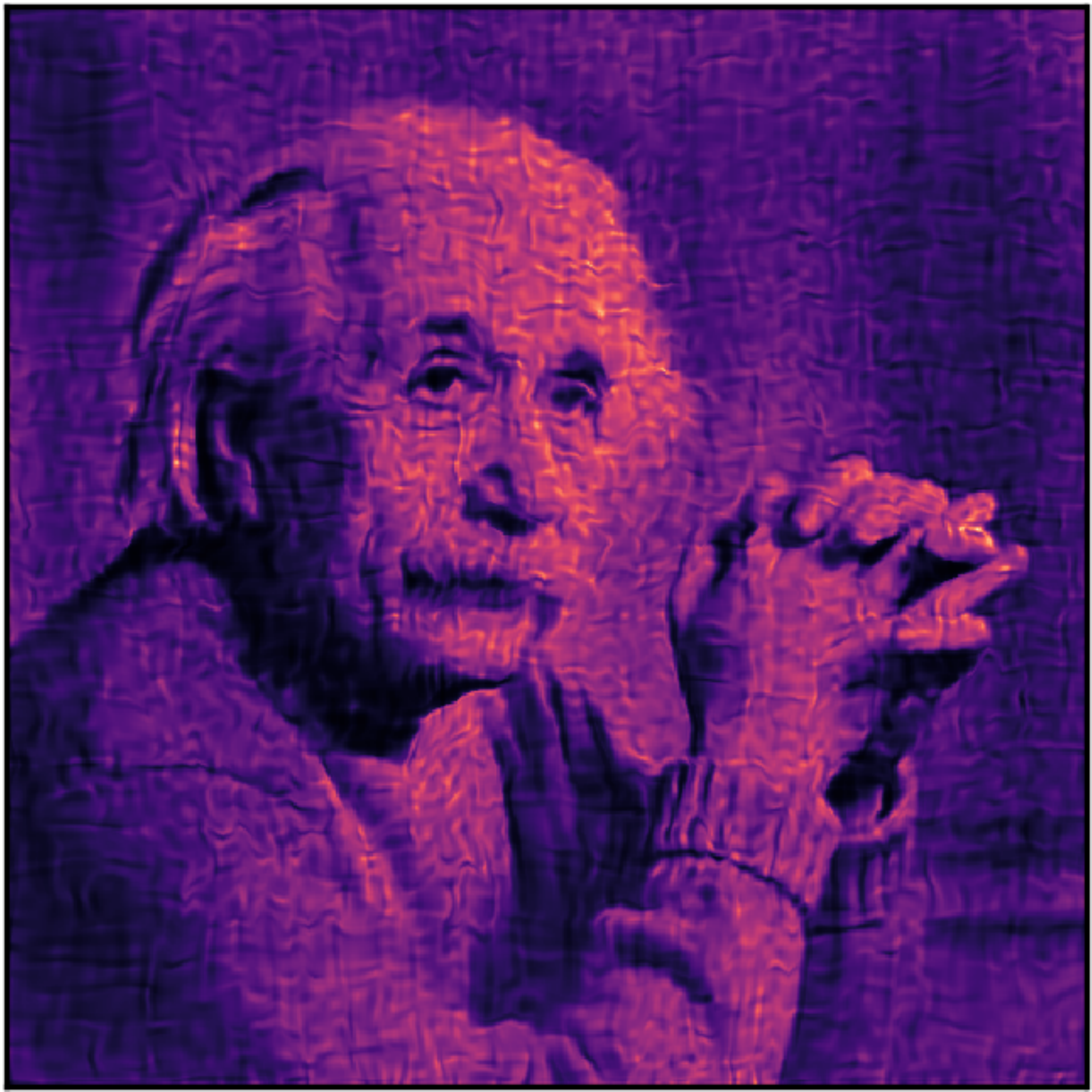} &
        	\includegraphics[width=\plotsizeplane]{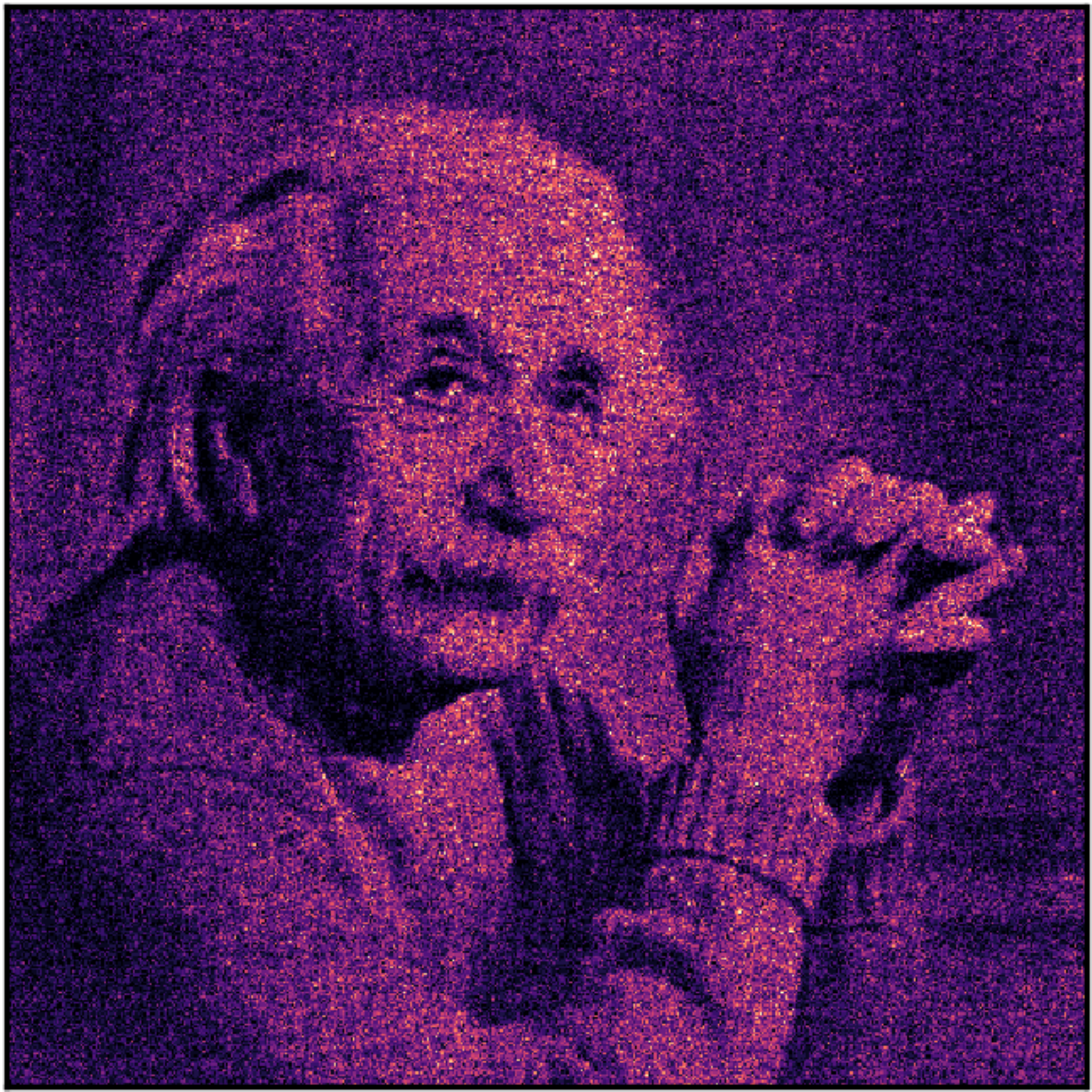}
        	\\
        	\includegraphics[width=\plotsizeplane]{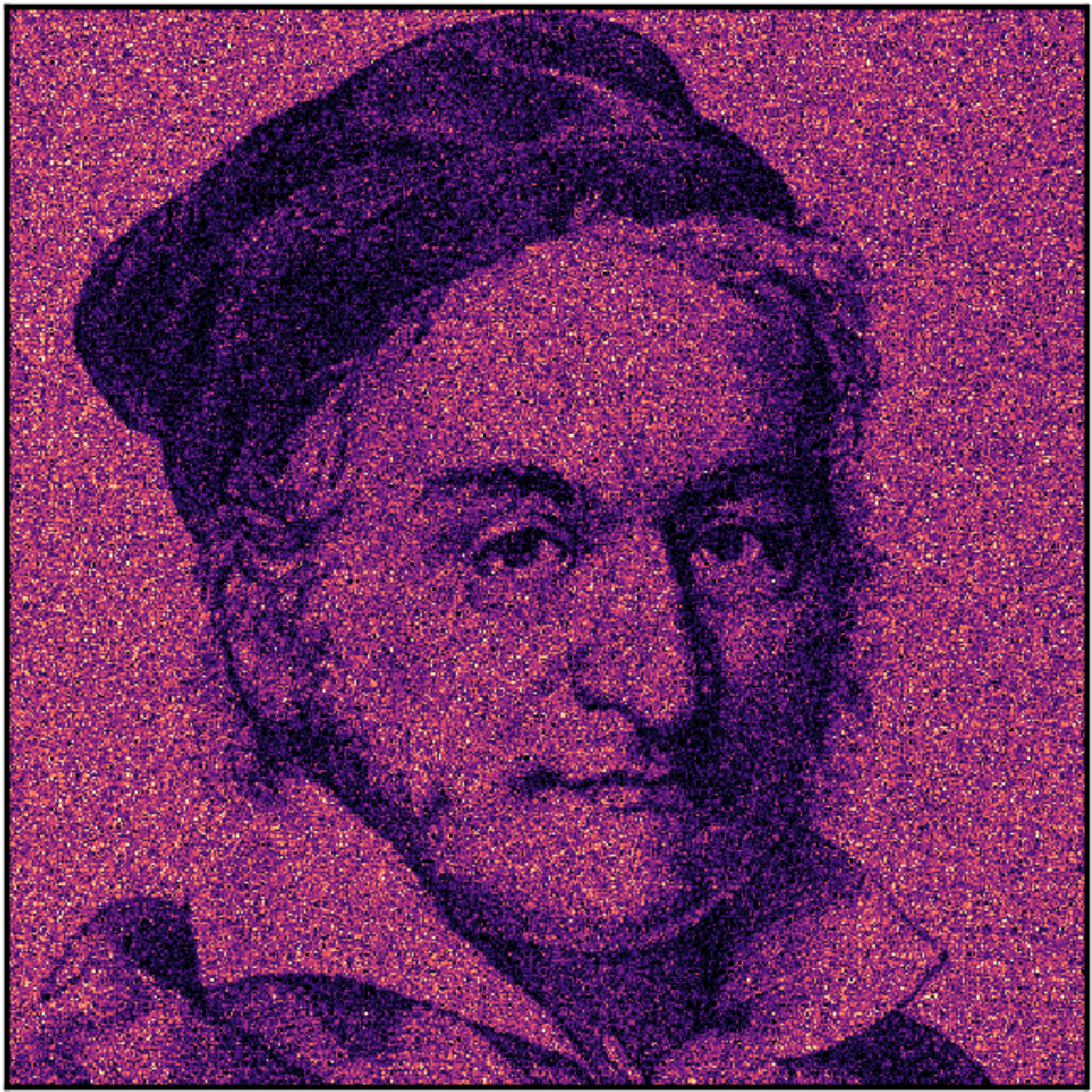} &
        	\includegraphics[width=\plotsizeplane]{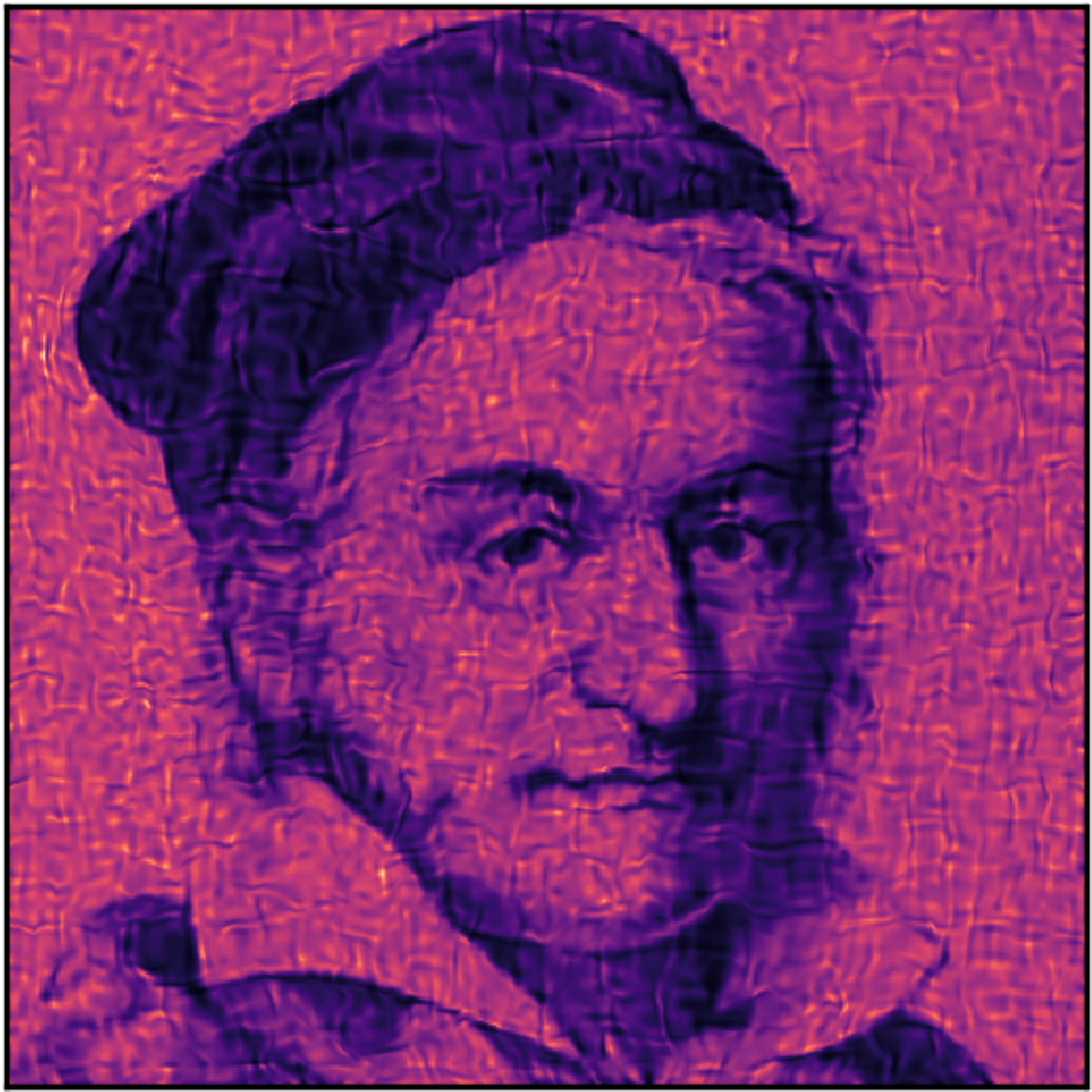} &
        	\includegraphics[width=\plotsizeplane]{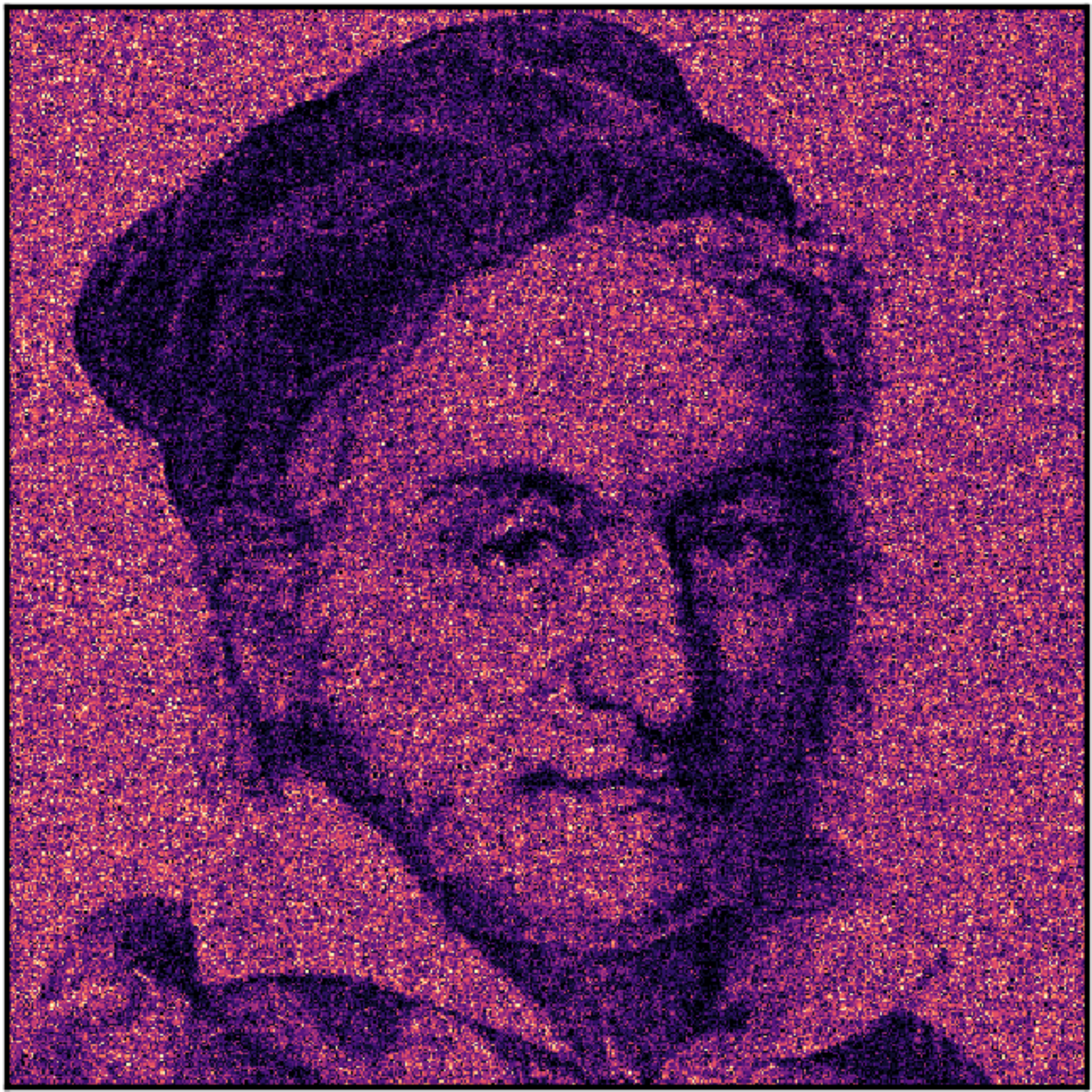}
        \end{tabular}
        \vspace{-0.6em}
        \caption{\small Qualitative results for two-dimensional synthetic datasets using cubic-spline flows with two coupling layers.
        Some previous flows struggle to model such fine details, as demonstrated by, e.g., \citet{nash2019aem}.}
        \vspace{-0.8em}
        \label{fig:plane-fits}
\end{figure}

\begin{table*}
	\caption{\small Test log likelihood (in nats) for UCI datasets and BSDS300; higher is better. Error bars correspond to two standard deviations (FFJORD do not report error bars).
        Apart from quadratic and cubic splines, all results are taken from existing literature. NAF$^\dagger$ report error bars across five repeated runs rather than across the test set.}
	\label{tab:uci-results}
        \vspace*{-0.1in}
	\begin{center}
		\begin{small}
			\begin{sc}
				\begin{tabular}{@{}llccccc@{}}
					\toprule
                    & Model                       & POWER           & GAS              & HEPMASS           & MINIBOONE         & BSDS300           \\ \midrule
                    \multirow{3}{*}{\begin{minipage}{6em}One-pass\\flows\end{minipage}} &
                    FFJORD {\rm \citep{Grathwohl:2018:ffjord}} & $0.46$\phantom{${}\pm 0.00$} & \phantom{$0$}$8.59$\phantom{${}\pm 0.00$} & $-14.92$\phantom{${}\pm 0.00$} & $-10.43$\phantom{${}\pm 0.00$} & $157.40$\phantom{${}\pm 0.00$}\\
                    & Quadratic-spline & $ 0.65 \pm 0.01$ & $ 13.13 \pm 0.02 $ & $ -14.95 \pm 0.02 $ & \phantom{$0$}$ -9.18 \pm 0.43 $ & $ 157.49 \pm 0.28 $ \\
                    & Cubic-spline & $0.65 \pm 0.01$ & $13.14 \pm 0.02$ & $-14.59 \pm 0.02$ & \phantom{$0$}$-9.06 \pm 0.44$ & $157.24 \pm 0.28$ \\ \midrule
                    \multirow{4}{*}{\begin{minipage}{6em}Auto-\\regressive\\flows\end{minipage}} &
                    MAF {\rm \citep{Papamakarios:2017:maf}} & $0.30 \pm 0.01$ & $10.08 \pm 0.02$ & $-17.39 \pm 0.02$ & $-11.68 \pm 0.44$ & $156.36 \pm 0.28$ \\
                    & NAF\smash{$^\dagger$} {\rm \citep{Huang:2018:naf}}& $0.62 \pm 0.01$ & $11.96 \pm 0.33$ & $-15.09 \pm 0.40$ & \phantom{$0$}$-8.86 \pm 0.15$ & $157.73 \pm 0.04$ \\
                    & Block-NAF \citep{DeCao:2019:bnaf} & $0.61 \pm 0.01$ & $12.06 \pm 0.09$ & $-14.71 \pm 0.38$ & \phantom{$0$}$-8.95 \pm 0.07$ & $157.36 \pm 0.03$ \\
                    & TAN various {\rm \citep{oliva2018}} & $0.60 \pm 0.01$ & $12.06 \pm 0.02$ & $-13.78 \pm 0.02$ & $-11.01 \pm 0.48$ & $159.80 \pm 0.07$ \\
                    \bottomrule
				\end{tabular}
			\end{sc}
		\end{small}
	\end{center}
\end{table*}

\section{Experiments}
For our experiments, NN in \cref{eq:coupling_nn} is a fully-connected or convolutional neural network with two pre-activation residual blocks \cite{he2016deep, he2016preact}. 
We use the Adam optimizer \cite{kingma2014adam}, and a cosine schedule for annealing the learning rate \cite{loshchilov2016sgdr}. 
All experiments use a standard normal for the noise distribution $\pi\roundbr{\bfu}$, except for the two-dimensional experiments, which use a uniform distribution.

\subsection{Synthetic datasets}
We first demonstrate the flexibility of cubic-spline flows on synthetic two-dimensional datasets (\cref{fig:plane-fits}). 
The  mixture of $225$ Gaussians and Einstein are taken from \citep{nash2019aem}, and we generate another dataset using an image of Gauss.
For each task, we train on one-million data points, use a cubic-spline flow with two coupling layers, no linear layers, and $ K = 128 $ bins.
As we can see, the model is flexible enough to fit complex, multimodal densities that other flows would struggle with. In each case, the model consists of fewer than $ 30\% $ of the parameters than the $ 512 \times 512 $ histograms which are used to display the data. 

\subsection{Density estimation on tabular data}
We next consider the UCI and BSDS300 datasets, a standard suite of benchmarks for density estimation of tabular data. 
Our spline flows alternate coupling layers with LU-decomposed linear layers, and we use $ K \!=\! 10 $ bins in all experiments. 
For all datasets, we stack ten of these composite linear and coupling layers, apart from MINIBOONE, where we use five. 
Hyperparameters such as the number of training steps and the hidden dimension used by NN in coupling layers are tuned separately for each cubic model.
As an ablation study, we compare with a flow using quadratic instead of cubic splines, which extends \cite{muller2018neuralimportancesampling} with LU layers and elementwise transformations as in \cref{sec:method}.

\Cref{tab:uci-results} shows our results. The cubic and quadratic models are close in performance, with the error bars on test log likelihood overlapping. However, a paired comparison computing the mean and standard error of the difference between models on individual examples shows that the cubic-spline flow is statistically significantly better on three of the five datasets, while being indistinguishable on POWER and slightly worse than the quadratic-spline flow on BSDS300. On the other hand, the quadratic-spline flow is slightly faster and less prone to numerical issues. \Cref{sec:cubic_vs_quadratic_spline} further compares the cubic and quadratic models.

Spline flows achieve state-of-the-art performance on all tasks against FFJORD \cite{Grathwohl:2018:ffjord}, the previously strongest flow-based model with a one-pass inverse. 
Moreover, the spline flows are competitive against the best-performing autoregressive flows, achieving state of the art for any flow model on POWER and GAS\@.
These results demonstrate that it is possible for neural density estimators based on coupling layers to challenge autoregressive models, and that it may not be necessary to sacrifice one-pass sampling for density-estimation performance.

\subsection{Image generation}
\begin{figure}[h]
	\centering
	\begin{subfigure}{0.26\columnwidth}
    	\centering
	    \includegraphics[width=\columnwidth]{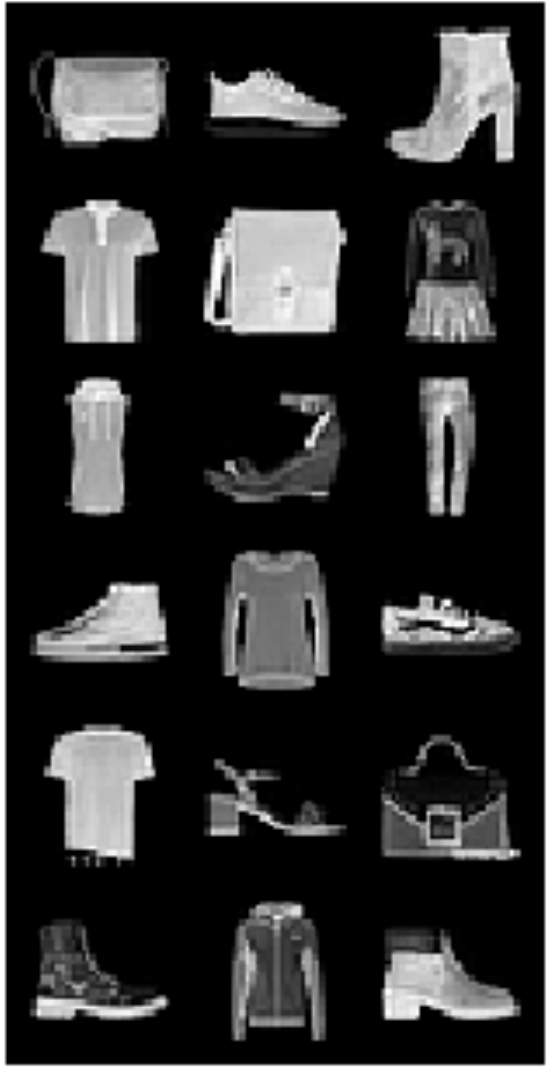}
	\end{subfigure}
	\begin{subfigure}{0.72\columnwidth}
	    \centering
	    \includegraphics[width=\columnwidth]{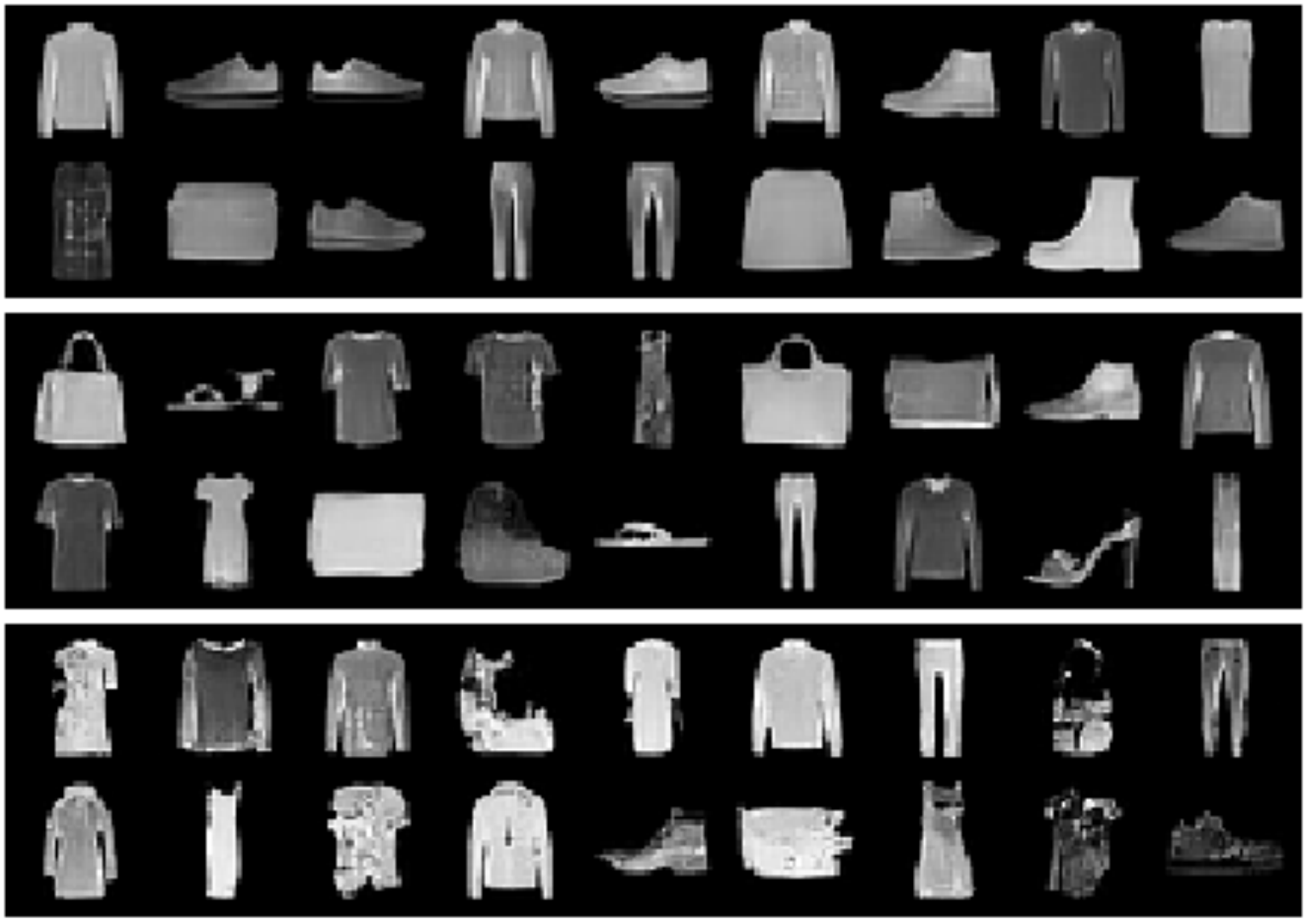}
	\end{subfigure}
	\caption{\small Qualitative results for Fashion-MNIST dataset. \textbf{Left}: Training data. \textbf{Right}: Unconditional samples from the cubic-spline flow. Samples for three temperatures are shown in separate blocks, top-to-bottom: $0.5, 0.75, 1.0$.}
	\label{fig:fashion-mnist}
\end{figure}

Finally, we demonstrate that cubic-spline flows scale to high-dimensional data using the Fashion-MNIST dataset \citep{xiao2017fashionmnist}. We use a Glow-like model \citep{Kingma:2018:glow} with cubic-spline coupling transforms, which includes a multi-scale architecture, actnorm layers with data-dependent initialization, and invertible $1\times1$ convolutions. In line with \citet{Kingma:2018:glow}, we illustrate the effect of scaling the standard deviation of the noise distribution $\pi\roundbr{\bfu}$ by a temperature $T \in [0,1]$. Qualitative results are shown in \cref{fig:fashion-mnist}. We leave the comparison with state-of-the-art flow-based image models on higher-dimensional image datasets for future work. 

\section{Conclusion}

The cubic-spline flow is the first model that combines the density-estimation performance of
state-of-the-art autoregressive models like NAF \citep{Huang:2018:naf} and TAN \citep{oliva2018}, while still being
able to generate realistic samples in a single pass like \mbox{Real NVP} \citep{Dinh:2017:rnvp}
and Glow \citep{Kingma:2018:glow}. More generally, we have shown that flexible
monotonic splines are a useful differentiable module, which could be included in
many models that can be trained end-to-end. We intend to implement
autoregressive flows with these transformations, which may further improve the
state of the art in density estimation.

\section*{Acknowledgements}
This work was supported in part by the EPSRC Centre for Doctoral Training in Data Science, funded by the UK Engineering and Physical Sciences Research Council (grant EP/L016427/1) and the University of Edinburgh. George Papamakarios was also supported by Microsoft Research through its PhD Scholarship Programme.

\appendix
\section{More details on monotonic cubic splines}

\subsection{Parameterization of the spline}
\label{sec:spline_parameterization}
We here include an outline of the method of \citet{steffen1990simple} which closely matches the original paper. 
Let $ \curlybr{\roundbr{x_{k}, y_{k}}}_{k=0}^{K} $ be a given set of knot points in the plane which satisfy
\begin{align}
&(x_0, y_0) = (0, 0), \quad (x_K, y_K) = (1, 1), \\
&x_{k} < x_{k+1} \,\text{ and }\, y_{k} < y_{k+1} \,\, \text{ for }k=0:K-1.
\end{align}
If additionally the derivatives $ \curlybr{d_{k}}_{k=0}^{K} $ at the knot points are known, a unique cubic polynomial
\begin{align}
    \label{eq:appendix:cubic-polynomial}
    f_{k}(\xi) = \alpha_{k0} + \alpha_{k1} \xi + \alpha_{k2} \xi^{2} + \alpha_{k3} \xi^{3},
\end{align}
where $ \xi = x - x_{k} $, is determined on each bin. 
The coefficients are given by
\begin{align}
    \alpha_{k0} &= y_{k} \\
    \alpha_{k1} &= d_{k} \\
    \alpha_{k2} &= \frac{3 s_{k} - 2 d_{k} - d_{k + 1}}{w_{k}} \\
    \alpha_{k3} &= \frac{d_{k} + d_{k + 1} - 2 s_{k}}{\roundbr{w_{k}}^{2}},
\end{align}
where $ w_{k} = x_{k + 1} - x_{k} $ is the width of each bin, and $ s_{k} = \roundbr{y_{k + 1} - y_{k}}/w_{k} $ is the slope of the line joining consecutive knots.
The resulting cubic spline is continuously differentiable on the interval $ [0, 1] $, and it remains only to determine the derivative values $ \curlybr{d_{k}}_{k=0}^{K} $ so that the overall spline is monotonic on this interval. 

To this end, \citet{steffen1990simple} uses the unique quadratic function passing through the points $ \roundbr{x_{k - 1}, y_{k - 1}}$, $\roundbr{x_{k}, y_{k}}$ and $\roundbr{x_{k + 1}, y_{k + 1}} $ to determine the derivative at location $ x_{k} $.
Monotonicity of this quadratic on the interval defined by these points is sufficient to specify an appropriate derivative, which is given by 
\begin{align}
    p_{k} = \frac{s_{k - 1} w_{k} + s_{k} w_{k - 1}}{w_{k - 1} + w_{k}}.
\end{align}
However, it is possible that the quadratic in question is not monotonic on the given interval, and in this case the derivative value needs to be modified.
\citet{steffen1990simple} proposes using the smallest value for the derivative at $ x_{k} $ such that the quadratic is monotonic on the given interval, showing that 
\begin{align}
   d_{k} = 
   \begin{cases}
   2 \min \roundbr{s_{k - 1}, s_{k}} \quad &\text{if } p_{k} > 2 \min \roundbr{s_{k - 1}, s_{k}} \\
   p_{k} \quad &\text{otherwise}
   \end{cases}
\end{align}
is sufficient.

\subsection{Computing the derivative}
\label{sec:spline_derivative}
The derivative of \cref{eq:appendix:cubic-polynomial} is straightforwardly given by
\begin{align}
    \label{eq:appendix:cubic-spline-derivative}
    \deriv{f_{k}(\xi)}{x} = \alpha_{k1} + 2 \alpha_{k2} \xi + 3 \alpha_{k3} \xi^{2}.
\end{align}
The logarithm of the absolute value of the determinant of a cubic-spline coupling transform is thus given by a sum of the logarithm of \cref{eq:appendix:cubic-spline-derivative} for each transformed $ x $.

\subsection{Computing the inverse}
\label{sec:spline_inverse}
Computing the inverse of a cubic polynomial requires solving for the roots of a cubic polynomial whose constant term depends on the value to invert. 
That is, for a given $ y $, we wish to find $ x $ which satisfies
\begin{align}
    \roundbr{\alpha_{k0} - y} + \alpha_{k1} \xi + \alpha_{k2} \xi^{2} + \alpha_{k3} \xi^{3} = 0,
\end{align}
where $ \xi $ is defined as before.
Monotonicity means there is at most one root $ x $ in each bin, and it is possible to determine this value analytically, using either Cardano's formula, or trigonometric or hyperbolic methods.
In practice, careful numerical consideration is necessary to ensure stable root-finding. 
We found the comprehensive treatment of \citet{blinn2007solve} particularly useful, ultimately settling on a slightly modified version of a procedure outlined in this latter work \citep{christoph2016howtosolve}. 
We refer readers to \citet{blinn2007solve} for full details. 

\subsection{Comparison with monotonic quadratic splines}
\label{sec:cubic_vs_quadratic_spline}
Following \citet{muller2018neuralimportancesampling}, we construct a quadratic spline by integrating an unconstrained linear spline. 
The result is a monotonic piecewise-quadratic function, where the values and derivatives of the function are continuous at the knots. 
The derivatives of the quadratic spline at the knots, corresponding to the density, are set directly by the linear spline. 
However, some monotonically-increasing knot locations $ \roundbr{x_{k}, y_{k}} $ cannot be expressed through this construction.

In contrast, Steffen's cubic-spline construction \citep{steffen1990simple} can interpolate arbitrary monotonic settings of the knot locations $ \roundbr{x_{k}, y_{k}} $, so unlike the quadratic spline can set arbitrary quantiles of the implied distribution at the knot locations. 
However, the method sets the derivatives of the function based on prior smoothness assumptions, rather than allowing the user to set them directly as in the quadratic spline. 
As a result, the number of parameters for the two spline constructions are similar.
The cubic spline could be made strictly more flexible by adding parameters for the derivatives at the knots, constrained to maintain monotonicity.

As they are, the cubic and quadratic splines give different functions for knots specifying the same quantiles, corresponding to different inductive biases. \citet{steffen1990simple} discusses several nice properties of the interpolants chosen by his cubic-spline method.

% Bibliography
\bibliographystyle{icml2019}
\bibliography{./bibliography.bib}

\end{document}